\documentclass{article}

\usepackage{arxiv}

\usepackage[utf8]{inputenc} 
\usepackage[T1]{fontenc}    
\usepackage{hyperref}       
\usepackage{url}            
\usepackage{booktabs}       
\usepackage{nicefrac}       
\usepackage{microtype}      
\usepackage{lipsum}
\usepackage{graphicx}
\graphicspath{{./figures/}}
\usepackage{hyperref}
\hypersetup{pdftex,colorlinks=true,allcolors=blue}
\usepackage{amsmath,amssymb,amsfonts}
\usepackage{stfloats}
\usepackage[]{algorithm2e}
\usepackage{siunitx}

\title{Assessment of Reward Functions for Reinforcement Learning Traffic Signal Control under Real-World Limitations \thanks{This paper has been accepted to the IEEE 2020 International Conference Systems, Man and Cybernetics (IEEE SMC2020) and was part funded by the EPSRC under grant no. EP/L015374 and part funded by InnovateUK grant no. 104219.}}

\author{
  Alvaro Cabrejas Egea \thanks{Alvaro Cabrejas Egea is with Mathematics for Real-World Systems Centre for Doctoral Training, University of Warwick.} \\
  MathSys Centre for Doctoral Training, \\
  University of Warwick \& Vivacity Labs\\
  CV4 7AL Coventry, NW1 2DB London, UK \\
  \texttt{a.cabrejas-egea@Warwick.ac.uk} \\
  \And
 Shaun Howell \thanks{Shaun Howell and Maksis Knutins are with Vivacity Labs, London.}\\
  Vivacity Labs\\
   NW5 3AQ London, UK \\
  \texttt{shaun@vivacitylabs.com} \\
  \And
Maksis Knutins $^\ddag$ \\
  Vivacity Labs\\
   NW5 3AQ London, UK \\
  \texttt{maksis@vivacitylabs.com} \\
   \And
 Colm Connaughton \thanks{Colm Connaughton is with Warwick Mathematics Institute, University of Warwick.}\\
  Warwick Mathematics Institute\\
  University of Warwick\\
  CV4 7AL Coventry, UK \\
  \texttt{c.p.connaughton@warwick.ac.uk} \\
}

\begin{document}
\maketitle

\begin{abstract}
Adaptive traffic signal control is one key avenue for mitigating the growing consequences of traffic congestion.
Incumbent solutions such as SCOOT and SCATS require regular and time-consuming calibration, can't optimise well for multiple road use modalities, and require the manual curation of many implementation plans.

A recent alternative to these approaches are deep reinforcement learning algorithms, in which an agent learns how to take the most appropriate action for a given state of the system.
This is guided by neural networks approximating a reward function that provides feedback to the agent regarding the performance of the actions taken, making it sensitive to the specific reward function chosen.
Several authors have surveyed the reward functions used in the literature, but attributing outcome differences to reward function choice across works is problematic as there are many uncontrolled differences, as well as different outcome metrics.

This paper compares the performance of agents using different reward functions in a simulation of a junction in Greater Manchester, UK, across various demand profiles, subject to real world constraints: realistic sensor inputs, controllers, calibrated demand, intergreen times and stage sequencing. 
The reward metrics considered are based on the time spent stopped, lost time, change in lost time, average speed, queue length, junction throughput and variations of these magnitudes.
The performance of these reward functions is compared in terms of total waiting time.
We find that speed maximisation resulted in the lowest average waiting times across all demand levels, displaying significantly better performance than other rewards previously introduced in the literature.
\end{abstract}

\section{Introduction}
Improving the operation of urban road networks can significantly reduce traffic congestion, as well as promote active travel and public transport within a city and reduce emissions.
Optimising the timings of adaptive traffic signals can play a large role in achieving this by making effective use of green lights in response to the demand levels on each approach to a junction and by promoting progression across multiple junctions.
In the UK, the main incumbent algorithms are 'Microprocessor Optimised Vehicle Actuation' (MOVA) \cite{MOVA} for isolated junctions, and 'Split, Cycle and Offset Optimisation technique' (SCOOT) \cite{SCOOT} for regions of  up to 30 signalised junctions.
However, these algorithms have been improved only incrementally since their initial development in the 1980s, and so do not take advantage of either modern traffic data sources, nor modern computational resources. Modern traffic data sources, such as vision-based above-ground detectors, are able to provide much richer information about the road network, for example by providing live occupancy, flow, class of vehicle (e.g. car, van, bus) and individual vehicles' positions and speeds.
These, combined with advances in GPU technology, have paved the way for real-time machine learning approaches in this field, with deep reinforcement learning showing promising results. 

Reinforcement learning (RL) has been investigated in recent years as a potential next step in urban traffic control (UTC) systems, demonstrating the potential to outperform even well-calibrated systems currently in use \cite{wei2019a}.
However, most of the work to date is not intended to be directly applied to the real world. As such, all observed works in the literature overlook operational limitations of this application of RL.
Further, there is a gap in the literature regarding the choice of reward function for such an RL system, which is a critical aspect.
This paper provides a robust comparison of reward functions for RL, in the context of a junction in Greater Manchester, UK, a simulation of which has been calibrated using extensive data from Vivacity Labs vision-based sensors.
This research is directly translatable to real-world applications of the technology, and has since been deployed to manage real traffic. 

The paper is structured as follows: Section \ref{lit} reviews earlier work in the field and enumerates different reward functions used in the literature.
Section \ref{problem} states the mathematical problem and the Reinforcement Learning theoretical background.
Section \ref{methods} describes the implementation and characteristics of the agents and environment.
Section \ref{rewards} describes and provides the analytic expressions of the different reward functions being tested.
Section \ref{exp_setup} gives detail on the training, selection and evaluation of the agents.
Section \ref{results} shows the results of the experiments in terms of average waiting time of the vehicles.

\section{Related Work} \label{lit}
Previous studies have considered RL for UTC without focusing specifically on the choice of reward functions.
Initial research was centered around proof-of-concept, with studies such as \cite{wiering2000} and \cite{abdulhai2003} advocating for its potential use, and the ability of Q-Learning to perform better \cite{abdulhai2010} \cite{prashanth2011} than traditional UTC methods such as MOVA \cite{MOVA}, SCOOT \cite{SCOOT} and SCATS \cite{SCATS}. 
Later research looked into neural networks as a function approximator to estimate the value of state-action pairs whilst addressing discretisation issues raised previously \cite{richter2007} \cite{araghi2013}.
Recent research makes use of deep RL to estimate the state-action values for each state \cite{mannion} \cite{vanderpol2016} \cite{genders2016} \cite{gao2017} \cite{wan2018} \cite{liang2018} \cite{zhou2019} or to learn a policy directly that maps states to actions \cite{richter2007} \cite{mousavi2017} \cite{genders2018} \cite{gendersthesis} \cite{aslani2019}.

A number of publications have compiled RL methods. In \cite{rlparam}, early table based methods are summarised; in \cite{mannion} reward functions in a multi-junction network (delay between actions, difference in delay between actions, minimisation and balancing of queues, and minimising stops) are compared, and in \cite{genders2018} three different state representations are compared, finding similar performance with each. As the outcome was found not to be sensitive to state representation, the present work keeps the state representation constant. More recent studies focus on different RL approaches to this problem \cite{yau}, and \cite{wei2019} considers state representation, reward function, action definition and model specific distinctions (online-offline, policy-value, tabular-function approximation) in a survey across the field, but without performing comparisons.
Previous work typically approximates in terms of road geometry, traffic demand, and operational constraints, creating models of intersections that preclude real-world applicability.

This points to a gap in the literature of directly comparing a broad range of reward functions, in a well-calibrated, geometrically-accurate simulation which accounts for real-world limitations (e.g. safety constraints): this is the topic of this paper.

\section{Problem Background and Definition} \label{problem}
\subsection{Reinforcement Learning}
\begin{figure}[thpb]
    \centering
    \includegraphics[width=0.5\linewidth]{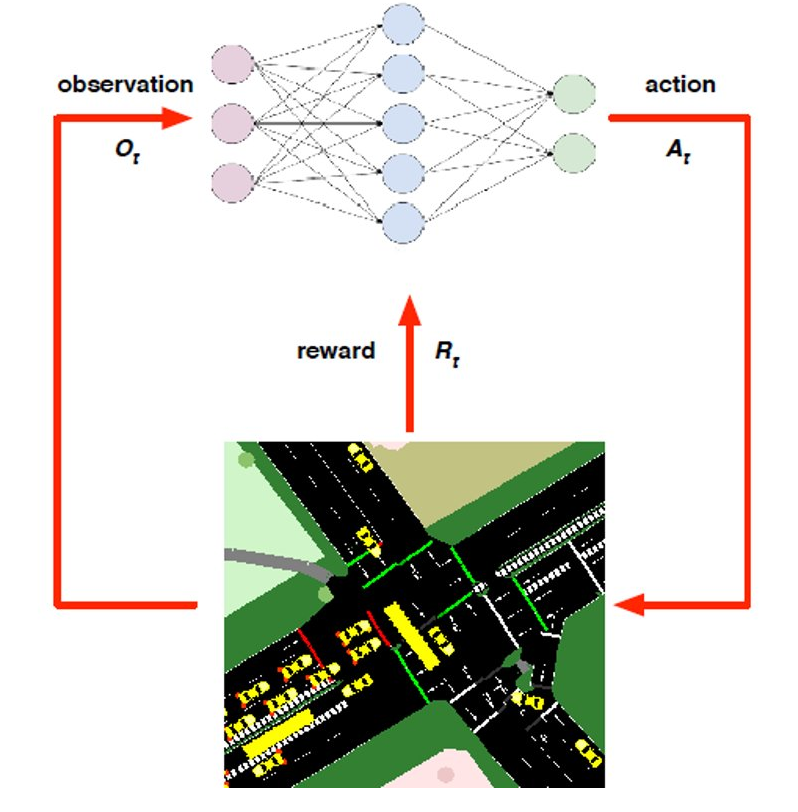}
    \caption{Schematic representation of information flows between Environment and Agent in a Reinforcement Learning framework.}
    \label{fig:RL}
\end{figure}

An intelligent agent is an entity which acts towards a goal based on observations and a decision making process. RL is an area of Machine Learning which focuses on how agents can learn a policy $\pi$ based on extended interaction with an environment.
This approach treats the problem as a Markov Decision Process (MDP), defined in terms of the  $\mathrm{<S, A, T, R, \gamma>}$ tuple, defined in accordance to what is shown in Fig. \ref{fig:RL}:
\begin{itemize}
    \item $\mathrm{S}$: Set of possible observable states. At time $t$, the agent will be provided with an observation $O_t=\{s^1_t, ..., s^i_t, ..., s^n_t\}$ of the state variables $s^i_t \in \mathrm{S}$.
    
    \item $\mathrm{A}$: Set of possible actions that the agent can execute in the environment. After being provided with $\mathrm{s_t}$, the agent chooses an action $a_t \in \mathrm{A}$.
    
    \item $\mathrm{T}$: Probabilistic state transition function.
    
    \item $\mathrm{R}$: Set of rewards that the agent will receive as a function of its performance. The reward $r_t(s,a)$ represents the reward at time $t$, obtained from performing action $a$ while being in state $s$.
    
    \item $\gamma \in [0,1]$: Discount factor for the reward of the next step, balancing the trade-off between future and immediate rewards.
\end{itemize}

The goal of the agent is to learn an optimal policy $\pi^*$ that maximises the expected future reward. The discounted future reward at time $t$, $R_t$, is defined in equation \eqref{eq:discounted_reward}.

\begin{equation}
R_t = \mathbb{E} \bigg{[} \sum^{\infty}_{i=0} \gamma^i r_{t+i} \bigg{]} 
\label{eq:discounted_reward}     
\end{equation}

\subsection{Q-Learning and Deep Q-Learning}
Q-Learning \cite{watkins} defines the value of a state action pair as the Q-Value $Q(s,a)$, which represents the value of taking a certain action $a$ while in state $s$, resulting in a transition to a new state $s'$.
$Q(s,a)$ is approximated by successive Bellman updates:
\begin{equation}
    Q_{t+1}(s,a) = Q_t(s,a) + \alpha [r + \gamma \max_{a'}Q_t(s',a')]
\label{eq:bellman}
\end{equation}

If $Q(s,a)$ is known, eq. (2) can be solved to obtain $\pi^*$ however its value is usually unknown, and so, estimating it is the task of Q-learning. Deep Q-Learning \cite{mnih2015}, thereby, uses a deep neural network as this function estimator.

\section{Methods} \label{methods}
\subsection{Agent}
The agent uses a Deep Q Network (DQN) implemented in PyTorch \cite{pytorch} with 2 fully connected hidden layers of 500 and 1000 neurons respectively, and an output layer of 2 neurons, one per allowed action.
All layers use ReLU as an activation function. The network weights are optimised using Stochastic Gradient Descent \cite{kiefer}, using ADAptive Moment Estimation (ADAM) \cite{adam} as the optimizer with a learning rate of $\alpha=10^{-5}$. The discount factor was set to $\gamma=0.8$ for all experiments.

\begin{algorithm}
\SetAlgoLined
 Initialise agent network with random parameters $\theta$\;
 Initialise target network with random parameters $\theta$'\;
 Initialise memory $M$ with capacity $L$\;
 Define frequency $F$ for copying weights to target network\;
 \For{each episode}{ 
    measure initial state $s_0$\;
    \While{episode not done}{
        choose action $a_t$ according to $\epsilon$-greedy policy\;
        implement action $a_t$\;
        advance simulator until next action needed\;
        measure new state $s_{t+1}$, and calculate reward $r_{t+1}$\;
        store transition tuple $(s_t,a_t,r_{t+1},s_{t+1})$ in $M$\;
        $s \leftarrow s_{t+1}$\;
        }
    $b \leftarrow$ sample batch of transitions tuples from $M$\;
    \For{each transition $x_i = (s_i,a_i,r_{i+1},s_{i+1})$ in $b$}{
        $y_i = r_{i+1} + \gamma \max_{a} Q(s_{i+1}, a', \theta')$
        }
    perform a step of Stochastic Gradient Descent on $\theta$ over all $(x_i,y_i) \in b$\;
    \If{number of episode is multiple of $F$}{
        $\theta' \leftarrow \theta$
        }
    }
 \caption{Schematic Learning Process}
\end{algorithm}
\subsection{Environment}

Our environment is a real four-arm junction located in Greater Manchester, UK.
This junction is modelled using the microscopic traffic simulator SUMO \cite{sumo} and calibrated using 3.5 months of flow and journey time data collected by vision-based sensors.
A stage is defined as a group of non-conflicting green lights (phases) in a junction.
The Agent decides which stage to select next and requests this from an emulated traffic signal controller, which moves to that stage subject to its limitations. These limitations are primarily safety-related and examples include enforcing minimum green times, minimum intergreen times, and stage transitions that match reality.
The site features four Vivacity vision-based sensors which can provide flow, queue length and speed data. The data available to the agent is restricted to what can be obtained from these sensors.
\begin{figure}[h]
    \centering
    \includegraphics[width=0.5\linewidth]{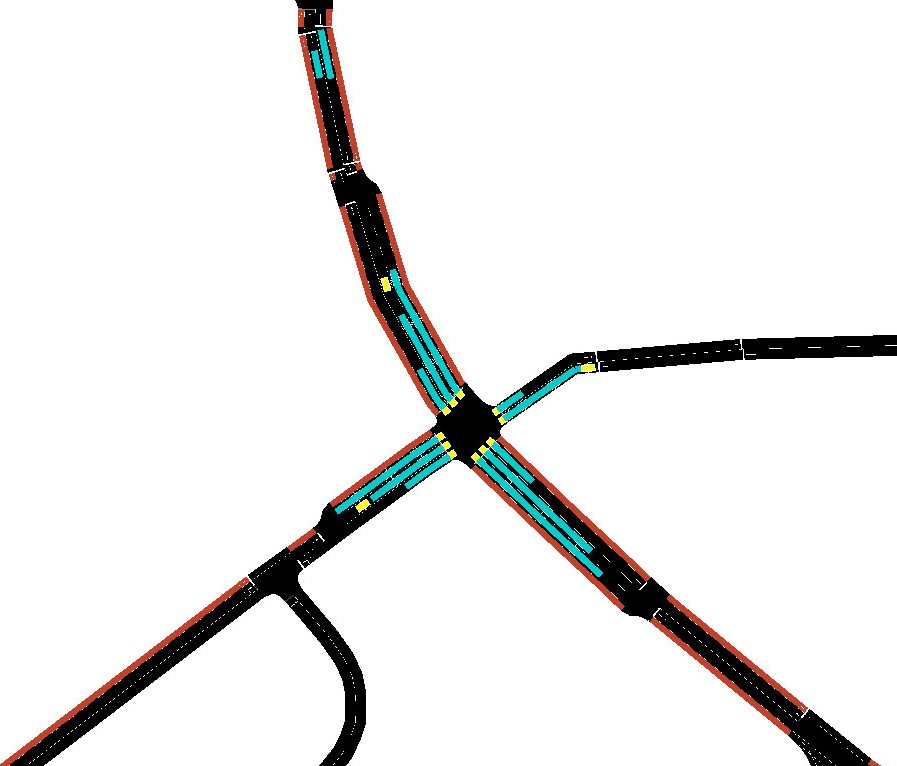}
    \caption{Intersection model in SUMO.}
    \label{fig:intersection}
\end{figure}

This paper does not consider pedestrians, thus promoting comparability with prior work; pedestrians will be considered for future work. 
\begin{figure}[thpb]
    \centering
    \includegraphics[width=0.5\linewidth]{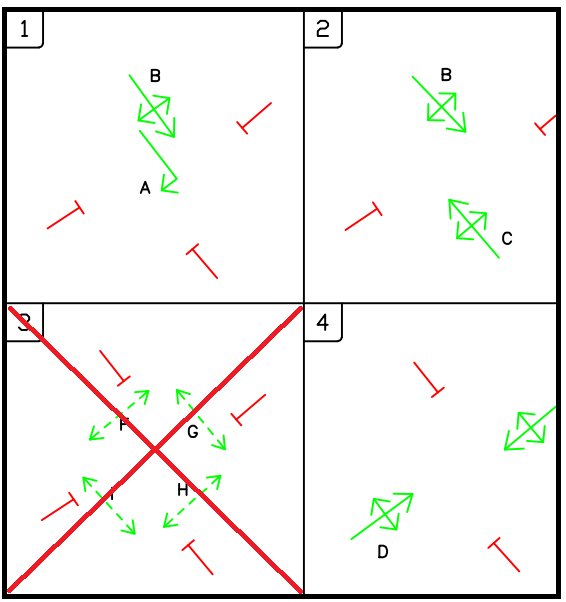}
    \caption{Allowed stages in the intersection (Stage 3 only serves pedestrians so isn't used). Stage 1 is an intermediate stage required to reach stage 2.}
    \label{fig:phases}
\end{figure}

\subsection{State Representation}
The input to the agent is a combination of two parts: sensor data and traffic controller state. 
The sensor data is the occupancy of each lane area detector, while the controller state is a one-hot representation of which stage is active.
A 12-second buffer at 0.6s resolution of both parts is provided to the agent.

While there are other state representations provided in the literature which are more information dense, many features of these cannot practically be obtained in the real world by the available sensors.
Moreover, recent findings \cite{genders2018} indicate that the gain from more information-dense states is marginal, meaning an Agent can manage an isolated intersection with relatively simple state inputs.
The state representation has been kept constant across the different experiments presented in this paper.

\subsection{Action Set}
The junction is configured to have 4 available stages. The Agent is able to choose Stage 2 or Stage 4, yielding an action space size of 2.
Stage 1 serves a leading right turn phase from the main road, and was excluded by suggestion of the transport authority, since it is an intermediary stage that the controller will go through in order to reach Stage 2, which serves the main road.
Stage 3 only serves pedestrians, which are not considered here, so was also excluded.
Stage 4 serves the side roads, which do experience significant demand.
In each timestep when a stage has been active longer than the minimum green time, the agent generates state-action values for each potential stage and the highest value is chosen according to an $\epsilon$-greedy policy \cite{suttonbarto}. If the agent chooses the same stage, that stage is extended by 0.6s, otherwise the controller begins the transition to the other stage.
The extension can be chosen indefinitely, as long as the agent identifies it as the best action.

The complexity in the decision-making stems from the combination of using Stage 1 as an intermediate state and the extensions to the stage duration.
Traditional RL for UTC regards each Stage as an action for the agent to take, based on the instantaneous state of the system.
However, in the case of the intermediate Stage 1, the agent has to choose when to start the transition without knowledge of the future state when Stage 2 begins.
Regarding the extensions, given that their length is smaller than that of the initial phase, their impact on the state will be smaller, generating a distribution of reward and state-action value outcomes that the agent needs to approximate.

\section{Reward Functions} \label{rewards}
In this section the individually tested reward functions are introduced.
Some of these have been obtained from the literature and are indicated as such, while some others are being first proposed here.
Any modifications to previously used functions are also stated.

Let $N$ be the set of lane queue sensors present in the intersection.
Let $V_t$ be the set of vehicles on incoming lanes in the intersection at time $t$, let $s$ be their individual speeds, $\tau$ their waiting times, and $\rho$ the flow across the intersection over the length of the action.
Let $t^p$ be the time at which the previous action was taken and $t^{pp}$ the time of the action before that.

\subsection{Queue Length based Rewards}
\subsubsection{Queue Length}
The reward will be the negative sum over all $n$ sensors of the queues ($q$) at time step $t$.
Similar to the reward introduced in \cite{prashanth2011} but without the need for thresholding the queue values, and used in \cite{aslani2019}.
\begin{equation}
    r_t = - \sum_{n \in N} q_{t}
\label{eq:queue}
\end{equation}
One of the first published in the field of Q-Learning, this reward function has low sensor requirements and is implementable using just induction loops.

\subsubsection{Queue Squared}
Introduced in \cite{gendersthesis}, this function squares the result of adding all queues.
This increasingly penalises actions that lead to longer queues.
\begin{equation}
   r_t = - \bigg( \sum_{n \in N} q_{t} \bigg)^2
\label{eq:queuesq} 
\end{equation}

\subsubsection{Delta Queue}
The reward will be the difference between the previous and current sum of queues, turning positive for those action that decrease the queue size and negative when it increases.
Similar approach to the rewards shown in Eqs. (\ref{eq:delta_wait_time}) and (\ref{eq:changedelay}).
\begin{equation}
    r_t = \sum_{n \in N} q_{t^p} - \sum_{n \in N} q_{t}
    \label{deltaqueue}
\end{equation}

\subsection{Waiting Time based Rewards}
\subsubsection{Wait Time}
The reward will be the negative aggregated time in queue ($\tau$) that the vehicles at the intersection have accumulated since the last action. 
\begin{equation}
r_t = - \sum_{v \in V_t} \tau_{t}
\label{eq:wait_time}
\end{equation}
This function is more information-dense than queues, scaling with individual waiting times, but requires more advanced hardware for individual vehicle recognition.

\subsubsection{Delta Wait Time}
Similar to Eq. \ref{deltaqueue}, used in \cite{liang2018}. The reward will be the difference between total intersection $\tau$ between the current time and the previous action. 
\begin{equation}
r_t =  \sum_{v \in V_t} \tau_{t_p} -  \sum_{v \in V_t} \tau_{t}
\label{eq:delta_wait_time}
\end{equation}

\subsubsection{Waiting Time Adjusted by Demand}
The reward will be the negative aggregated waiting time as above, but in this case it is divided by an estimate of the current demand ($\hat{d}$) or arrival rate, implicitly accepting that given a wait time as a result of an action, the penalty should scale with the difficulty of the task.
\begin{equation}
   r_t = -\frac{1}{\hat{d}} \sum_{v \in V_t} \tau_{t}
\label{eq:wait_time_norm} 
\end{equation}

\subsection{Time Lost based Rewards}
\subsubsection{Time Lost}
Used in \cite{wan2018}, the reward will be the negative aggregated delay accumulated by all vehicles upstream from the intersection, understanding the delay as deviations from the vehicle's maximum allowed speed ($s_{max}$). 
Assuming a simulator time step of length $\delta$:
\begin{equation}
    r_t = - \sum_{v \in V_t}  \sum_{t^p}^t \delta \big( 1-\frac{s_v}{s_{max}} \big)
\label{eq:delay}
\end{equation}
This reward provides a more accurate representation of the total delay caused, since it also accounts for all deceleration happening around the intersection. 

\subsubsection{Delta Time Lost}
Introduced in \cite{abdulhai2010} and used in \cite{mannion} \cite{genders2016} \cite{gao2017} \cite{mousavi2017} and \cite{genders2018}.
Similar to Eq. (\ref{eq:delta_wait_time}).
The reward will be the change of global delay in the vehicles around the intersection since the last action was taken.
\begin{equation}
   r_t = \sum_{v \in V_t}  \sum_{t^{pp}}^{t^p} \delta  \big( 1-\frac{s_v}{s_{max}} \big) - \sum_{v \in V_t}  \sum_{t^p}^t \delta  \big( 1-\frac{s_v}{s_{max}} \big)
\label{eq:changedelay} 
\end{equation}
This reward function provides both punishment and reward centered around zero.

\subsubsection{Delay Adjusted by Demand}
The reward will be the same as in the point above, but divided by an estimate of the demand level ($\hat{d}$).
\begin{equation}
    r_t = -\frac{1}{\hat{d}} \sum_{v\in V_t}  \sum_{t^p}^t \delta  \big( 1-\frac{s_v}{s_{max}} \big)
\label{eq:delay_dn}
\end{equation}

\subsection{Average Speed based Rewards}
\subsubsection{Average Speed}
This reward seeks to maximise the average joint speed of all vehicles around an area of influence around the intersection.
\begin{equation}
    r_t = \frac{1}{\sum_{V_t} v} \sum_{v \in V_t} \big( \frac{s_v}{s_{max}} \big)
\label{eq:avgspeed}
\end{equation}

\subsubsection{Average Speed Adjusted by Demand}
The reward will be, as in the previous section, but multiplied by an estimation of the demand ($\hat{d}$). This function scales the reward with the difficulty of the task.
\begin{equation}
    r_t = \frac{\hat{d}}{\sum_{V_t} v} \sum_{v \in V_t} \big( \frac{s_v}{s_{max}} \big)
\label{eq:avgspeed_dn}
\end{equation}

\subsection{Throughput based Rewards}
\subsubsection{Throughput}
The reward will be the total number of vehicles that cleared the intersection between the last time that an action was taken and now.
\begin{equation}
    r_t = \sum_{t_p}^t \rho
\label{eq:throughput}
\end{equation}

\subsubsection{Other reward functions}
There are several other reward functions which were not considered. For example minimising the frequency of signal change \cite{vanderpol2016} \cite{wei2018}, and accident avoidance \cite{vanderpol2016}: both of these concerns are already addressed by traffic signal controllers. Also, pressure has been used as a reward function in the control of a large network \cite{wei2019a} \cite{yau} \cite{wei2019} \cite{varaiya2013} \cite{wei2018}  \cite{chen2020} and achieved good results; however, this requires data from upstream and downstream of the target junction, so is beyond the current scope.

\section{Experimental Setup}\label{exp_setup}
\subsection{Training Process}

In each training run, the agent is subject to a training curriculum including a variety of scenarios, including sub-saturated, near-saturated, and over-saturated situations. The agent is first shown sub-saturated episodes, before the difficulty level is increased. Each episode runs for 3000 steps of length 0.6 seconds, for a total simulated time of 30 minutes (1800 seconds).

Ten training runs were conducted for each of the reward functions described in the following section, to capture variance in training run outcome.

\subsection{Evaluation and Scoring}
The performance assessment is done in terms of waiting time, a common metric used in traffic signal optimisation.
After each training run is complete, the agent's performance is evaluated and compared to that of reference agents. Two reference agents have been implemented to give context to the RL performance: Maximum Occupancy (longest queue first) and a Vehicle Actuated Controller. The Vehicle Actuated algorithm was an implementation of System D \cite{highways}, a common algorithm in the UK.  For each training run, the RL and reference agents are each tested on 300 scenarios: this is 100 scenarios at each of 3 demand levels. In each repetition the average waiting (stopped) time for all vehicles was computed.

The first scenario involves a demand of 1714 vehicles/hour (1 vehicle/2.1 seconds), and will be referred as Low Demand scenario.
The second scenario uses a demand of 2117 vehicles/hour (1 vehicle/1.7 seconds), and will be referred as Medium Demand scenario, although this level of demand is around the observed peak and the saturation level for the junction.
The third scenario test a demand level slightly above what the junction can currently serve. It uses a demand of 2400 vehicles/hour (1 vehicle/1.5 seconds) and will be referred as High Demand scenario.

For each reward function, the best result of the 10 training runs was selected, based on the above evaluation method.

\section{Results}\label{results}
In this section, the distribution of average waiting times per vehicle across the different scenarios is presented for each reward function.

\begin{figure}[thpb]
    \centering
    \includegraphics[width=0.7\linewidth]{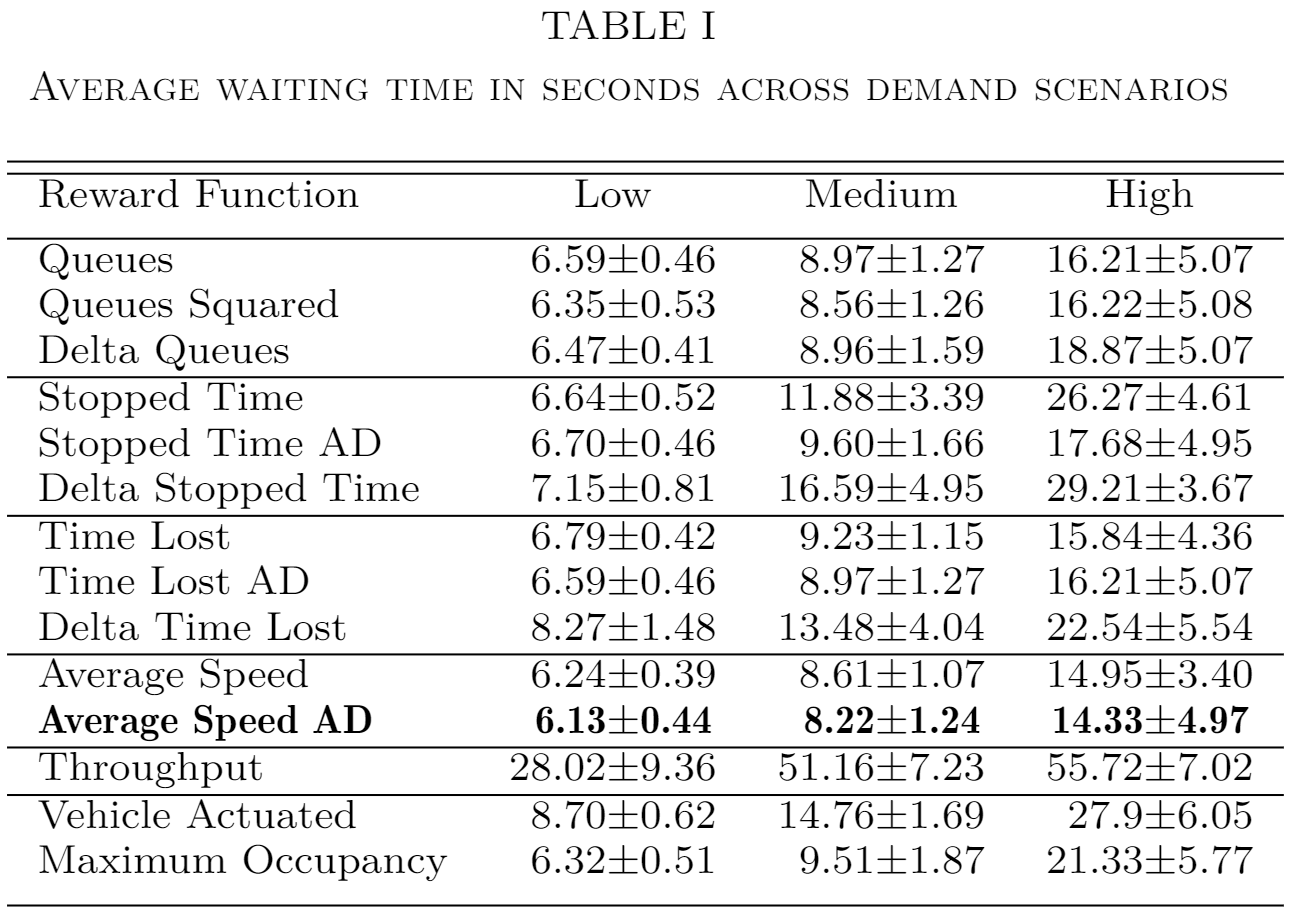}
    \label{fig:table}
\end{figure}

\begin{figure}[thpb]
    \centering
    \includegraphics[width=0.7\linewidth]{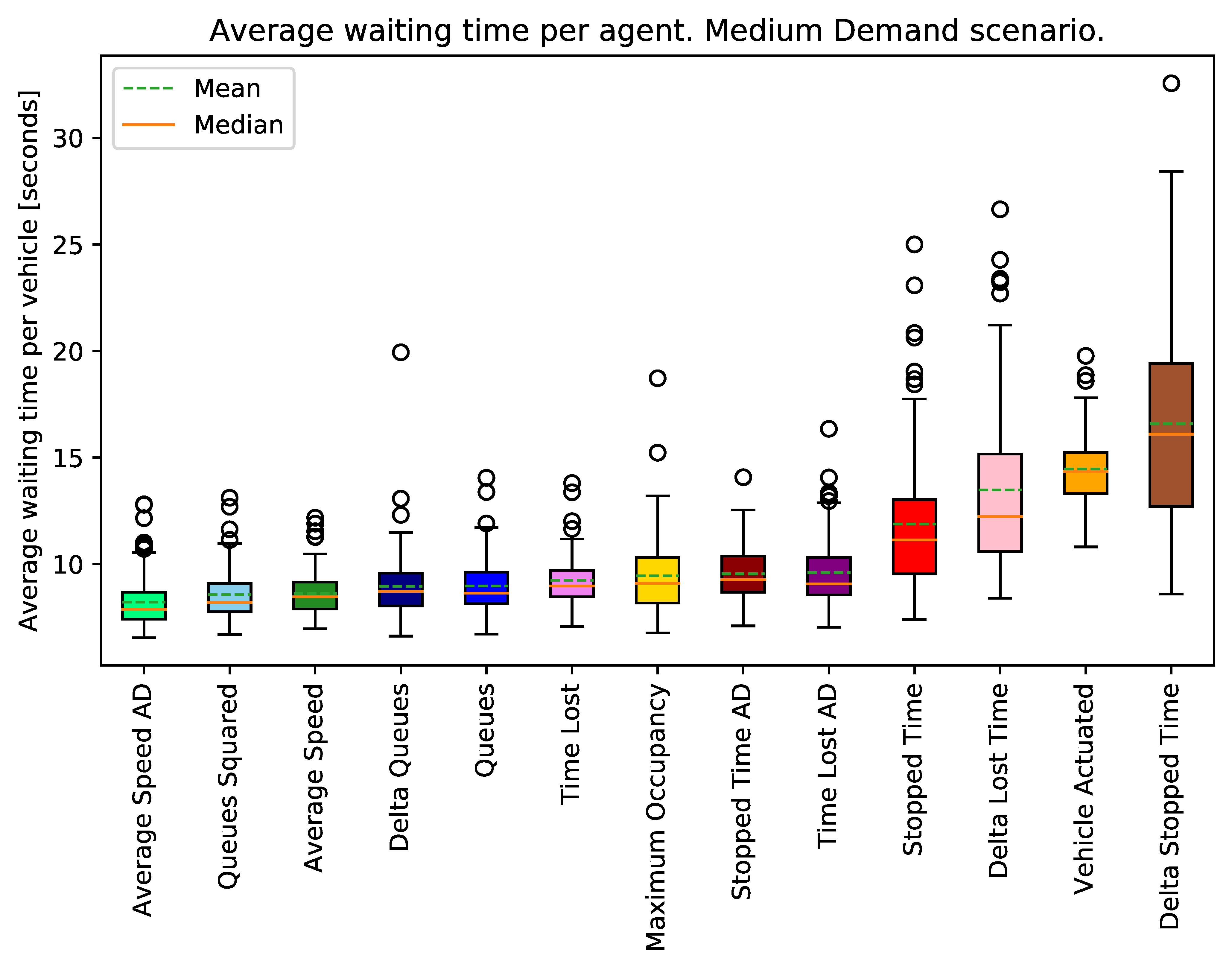}
    \caption{Distribution and medians of average waiting time in seconds across agents in Low Demand scenario. Sub-saturation demand of 1714 vehicles/hour (1 vehicle/2.1 seconds).}
    \label{fig:low}
\end{figure}

\begin{figure}[thpb]
    \centering
    \includegraphics[width=0.7\linewidth]{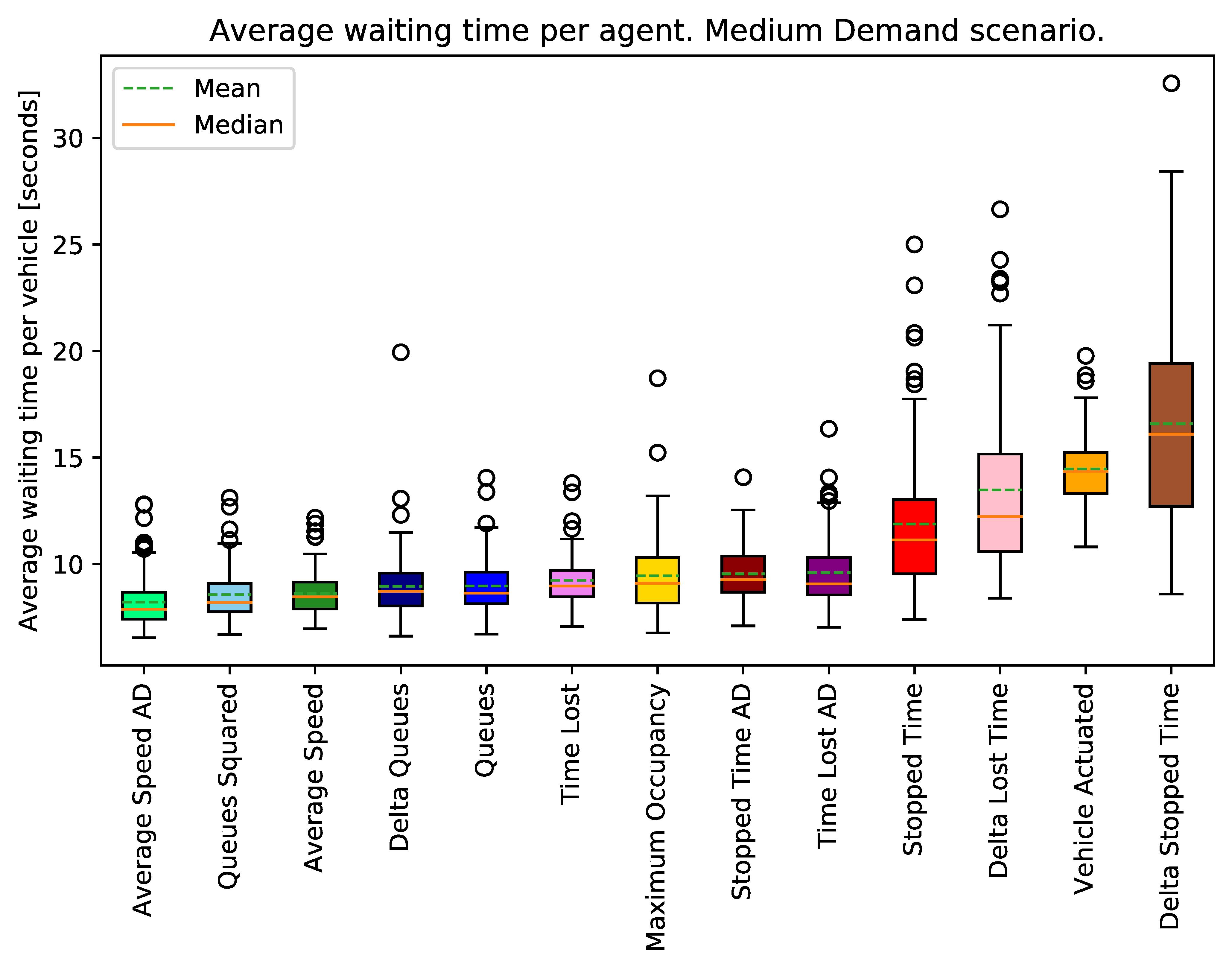}
    \caption{Distribution and medians of average waiting time in seconds across agents in Medium Demand scenario. Near-saturation demand of 2117 vehicles/hour (1 vehicle/1.7 seconds).}
    \label{fig:mid}
\end{figure}

\begin{figure}[thpb]
    \centering
    \includegraphics[width=0.7\linewidth]{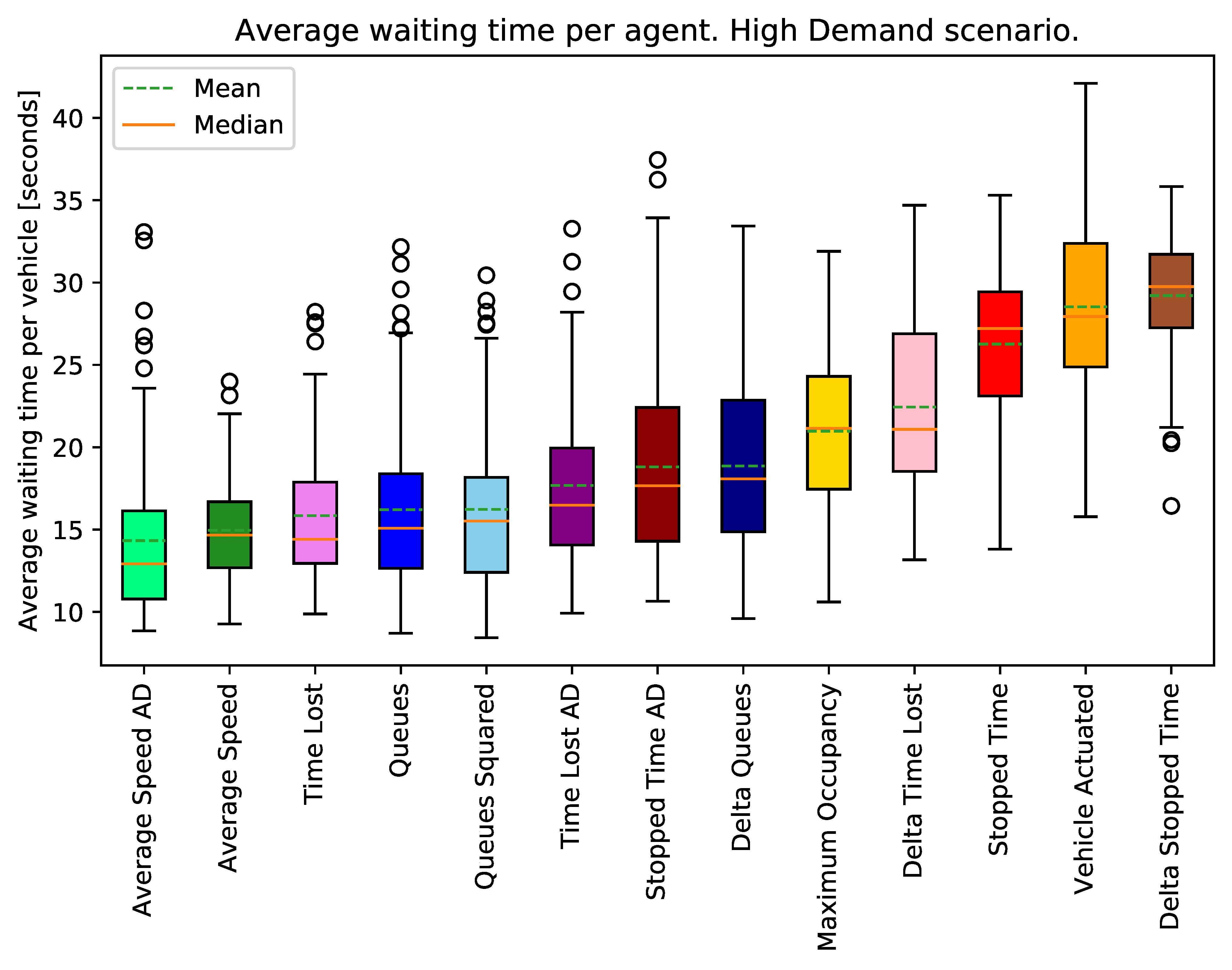}
    \caption{Distribution and medians of average waiting time in seconds across agents in High Demand scenario. Over-saturation demand of 2400 vehicles/hour (1 vehicle/1.5 seconds).}
    \label{fig:high}
\end{figure}

Figures \ref{fig:low} - \ref{fig:high} show for each agent, the distribution of average waiting times per vehicle across 100 repetitions of each scenario as described in the section above.
The throughput reward function is not shown to improve graph readability, as it performed poorly by all metrics. 
Complete corresponding results are reported in Table I.

\begin{figure}[thpb]
    \centering
    \includegraphics[width=0.7\linewidth]{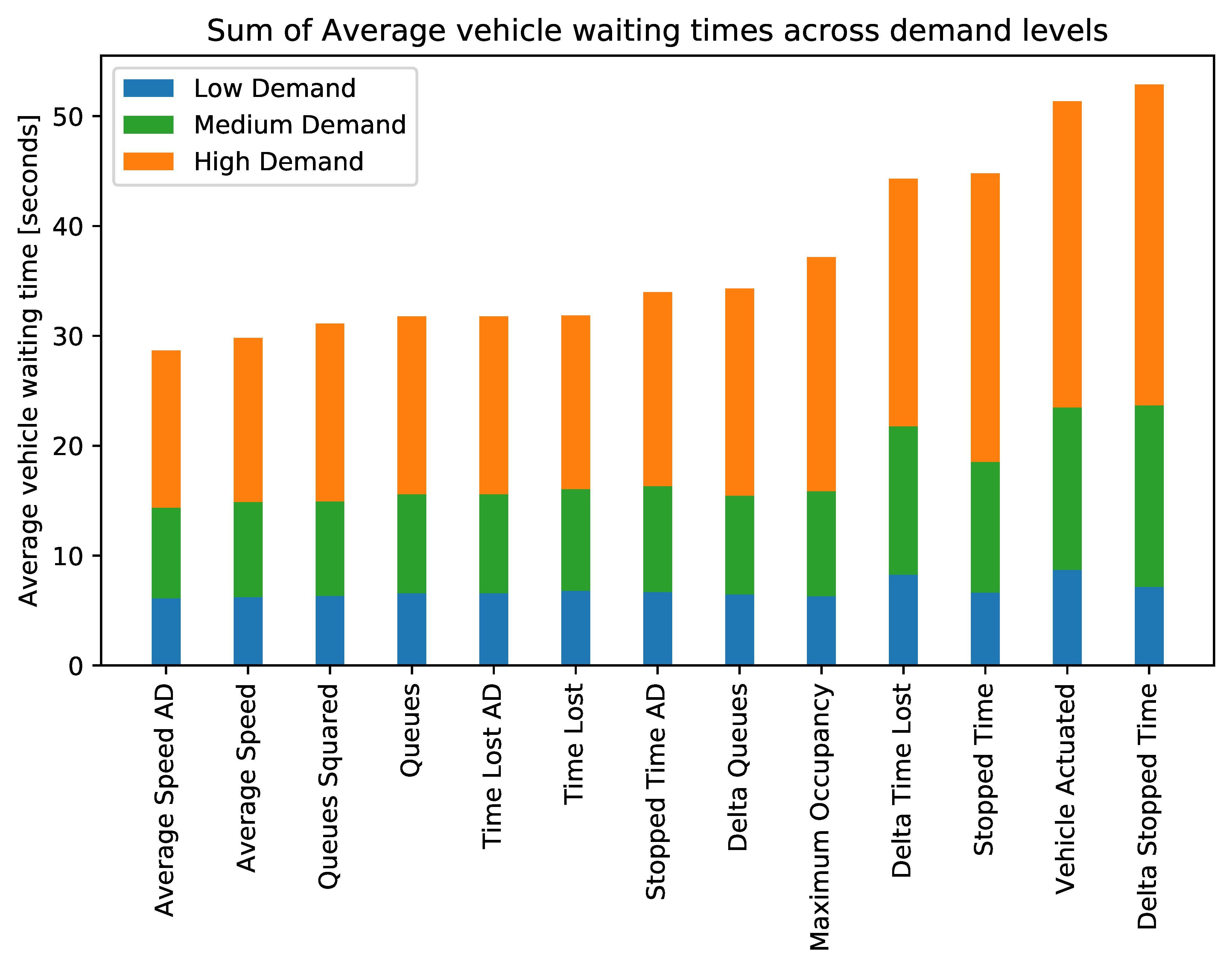}
    \caption{Stacked bar chart of Average Waiting Time across scenarios.}
    \label{fig:stack}
\end{figure}

Reward functions optimising the Average Speed of around the intersection were found to be best performing across all three scenarios and consistently outperformed the reference agents.
Adjusting the average speed by an estimate of the demand level shows to improve the performance of this reward function in all cases.

From those rewards using queues, Queues Squared performed best, except for the High Demand scenario in which the sum of Queues obtains marginally lower waiting times.

The results from agents using reward functions based on either Wait Time or Time Lost are inconsistent across different demand levels.
While sum of Time Lost is the third best-performing agent at High Demand, it is sixth in Medium Demand and Tenth in Low Demand.
Sum of Wait Time follows an inverse, yet less extreme trajectory as the demand is lowered, from worse to better performance.
Throughput was found to be consistently the worst-performing of the reward functions tested.
The rewards based on difference in delay (Delta Wait Time and Delta Time Lost), which were used in seven previous items of research as a most desirable reward function, were amongst the worst performers.


\section{Discussion and Conclusion}
This paper has explored the performance of several reward functions for deep Q-learning agents in the context of Urban Traffic Control. The paper reaffirms earlier findings that RL outperforms deterministic baseline algorithms.
Overall, the proposed reward functions based on maximisation of the average speed of the vehicles in the network resulted in the lowest waiting times across demand levels, including both baseline algorithms.
The performance of queue length and stopped time rewards was comparable, but when reliability is taken into account, average speed is clearly preferred since it produces a smaller variation in waiting times.

In sub-saturated conditions, a simple approach of always serving the largest queue probably suffices, although rewarding based on queue length had similar performance, and rewarding for average speed yielded slight improvements.
In busier conditions, which are more important to transport authorities, RL provides significant improvements in waiting time, with speed-based rewards again performing best, followed by queue-based rewards in near-saturation conditions, and followed by penalising for lost time in saturated conditions.

Rewarding based on delay minimisation or stopped time may perform worse than speed-based rewards as they encode information about previous timesteps, whereas queue lengths and vehicle speeds are snapshots at a single timestep, which better suits the underpinning Markov Decision Process framing of the problem.
Speed maximisation may have yielded better results than queue minimisation under high demand as it penalises the agent for vehicles which are moving slowly forward in congested conditions, which would not have been penalised by the queue length reward.
This would also not have been captured by the stopped time reward, which may explain its worse performance overall, even while being the target metric. One key finding is that whilst rewards based on difference in delay were found to perform well in earlier works, this was found to perform poorly in the present work.
The Maximum Occupancy reference agent performs well at low demand, but mediocre at medium demand and poorly at high demand.
It does not extend stage durations in busy traffic when the density of vehicles using the active stage is slightly lower than that of stationary queuing traffic.
This results in shorter than optimal stage lengths and so too much time is wasted in interstage transitions.
However, it did consistently perform better than the Vehicle Actuated reference agent.
The Vehicle Actuated algorithm extends stages until the related loops have been unoccupied for 1.5s, hence extending stages in congested traffic.
However, its performance suggests that this standard extension length, coupled with the maximum stage durations found empirically to perform relatively well, is overly eager to extend stages.
From a visual inspection via the simulator of the best RL agent's behaviour, it appears to learn an 'adaptive stage extension' behaviour: learning when to prioritise a large queue waiting to be served and when to prioritise avoiding the cost of transitioning stages.
A key benefit to the DQN approach is that it can learn how eagerly it should extend stages under varying conditions at a specific site, rather than requiring manual calibration.
This automated calibration can be repeated regularly so that the site continues to perform well long-term without manual intervention.

The methodology presented in this paper can be easily translated to other intersections.
The main requirements are a topologically accurate model of the target intersection, a realistic demand distribution, a set of sensors reporting the required state variables of the system, and a set of allowed actions on which to train the agent.

One limitation of the study is that the results presented only relate to the training of a deep Q-learning agent as specified in Section \ref{methods}, and are not necessarily representative of the performance in other type of architectures such as those representing the state as an image and processing it with Convolutional Neural Networks, or Policy Gradient methods.
Another limitation is that most intersections found in an urban setting do not exclusively serve vehicles. As stated in Section \ref{methods}, a similar exploration including pedestrians is left as future work for modal prioritisation based rewards.

\section*{Acknowledgment}
A.C.E thanks N. Walton and W. Chernicoff for the insights and discussions that led to this paper, and the Toyota Mobility Foundation for previous support that made this work possible. Vivacity Labs thanks Transport for Greater Manchester for helping take this work from simulation to the real world, Immense for their work calibrating the simulation model, and InnovateUK for funding that made the work possible.


\end{document}